\def\year{2019}\relax
\documentclass[letterpaper]{article} 
\usepackage{aaai19}  
\usepackage{times}  
\usepackage{helvet}  
\usepackage{courier}  
\usepackage{url}  
\usepackage{graphicx}  
\usepackage{subcaption} 
\usepackage{algorithm}
\usepackage{algorithmic}
\usepackage{booktabs}
\usepackage{multirow}
\usepackage{amsmath}
\usepackage{amssymb}

\usepackage{array}
\usepackage{bbm}

\newcommand{\citet}[1]
{\citeauthor{#1}˜\shortcite{#1}}
\newcommand{\citep}{\cite}

\title{CheXpert: A Large Chest Radiograph Dataset \\ with Uncertainty Labels and Expert Comparison}
\author{
Jeremy Irvin,\textsuperscript{1,*}
Pranav Rajpurkar,\textsuperscript{1,*}
Michael Ko,\textsuperscript{1}
Yifan Yu,\textsuperscript{1}\\
\bf \Large Silviana Ciurea-Ilcus,\textsuperscript{1}
\bf \Large Chris Chute,\textsuperscript{1}
\bf \Large Henrik Marklund,\textsuperscript{1}
\bf \Large Behzad Haghgoo,\textsuperscript{1}\\
\bf \Large Robyn Ball,\textsuperscript{2}
\bf \Large Katie Shpanskaya,\textsuperscript{3}
\bf \Large Jayne Seekins,\textsuperscript{3}
\bf \Large David A. Mong,\textsuperscript{3}\\
\bf \Large Safwan S. Halabi,\textsuperscript{3}
\bf \Large Jesse K. Sandberg,\textsuperscript{3}
\bf \Large Ricky Jones,\textsuperscript{3}
\bf \Large David B. Larson,\textsuperscript{3}\\
\bf \Large Curtis P. Langlotz,\textsuperscript{3}
\bf \Large Bhavik N. Patel,\textsuperscript{3}
\bf \Large Matthew P. Lungren,\textsuperscript{3,\dag}
\bf \Large Andrew Y. Ng\textsuperscript{1,\dag}\\
\textsuperscript{1}{Department of Computer Science, Stanford University}\\
\textsuperscript{2}{Department of Medicine, Stanford University}\\
\textsuperscript{3}{Department of Radiology, Stanford University}\\
\textsuperscript{*}{Equal contribution}\\
\textsuperscript{\dag}{Equal contribution}\\
\{jirvin16, pranavsr\}@cs.stanford.edu
}

\begin{document}
\maketitle

\begin{abstract} 
Large, labeled datasets have driven deep learning methods to achieve expert-level performance on a variety of medical imaging tasks. We present CheXpert, a large dataset that contains 224,316 chest radiographs of 65,240 patients. We design a labeler to automatically detect the presence of 14 observations in radiology reports, capturing uncertainties inherent in radiograph interpretation. We investigate different approaches to using the uncertainty labels for training convolutional neural networks that output the probability of these observations given the available frontal and lateral radiographs. On a validation set of 200 chest radiographic studies which were manually annotated by 3 board-certified radiologists, we find that different uncertainty approaches are useful for different pathologies. We then evaluate our best model on a test set composed of 500 chest radiographic studies annotated by a consensus of 5 board-certified radiologists, and compare the performance of our model to that of 3 additional radiologists in the detection of 5 selected pathologies. On Cardiomegaly, Edema, and Pleural Effusion, the model ROC and PR curves lie above all 3 radiologist operating points. We release the dataset to the public as a standard benchmark to evaluate performance of chest radiograph interpretation models.\footnote{https://stanfordmlgroup.github.io/competitions/chexpert}
\end{abstract}

\begin{figure}[t]
\centering
\includegraphics[width=\columnwidth]{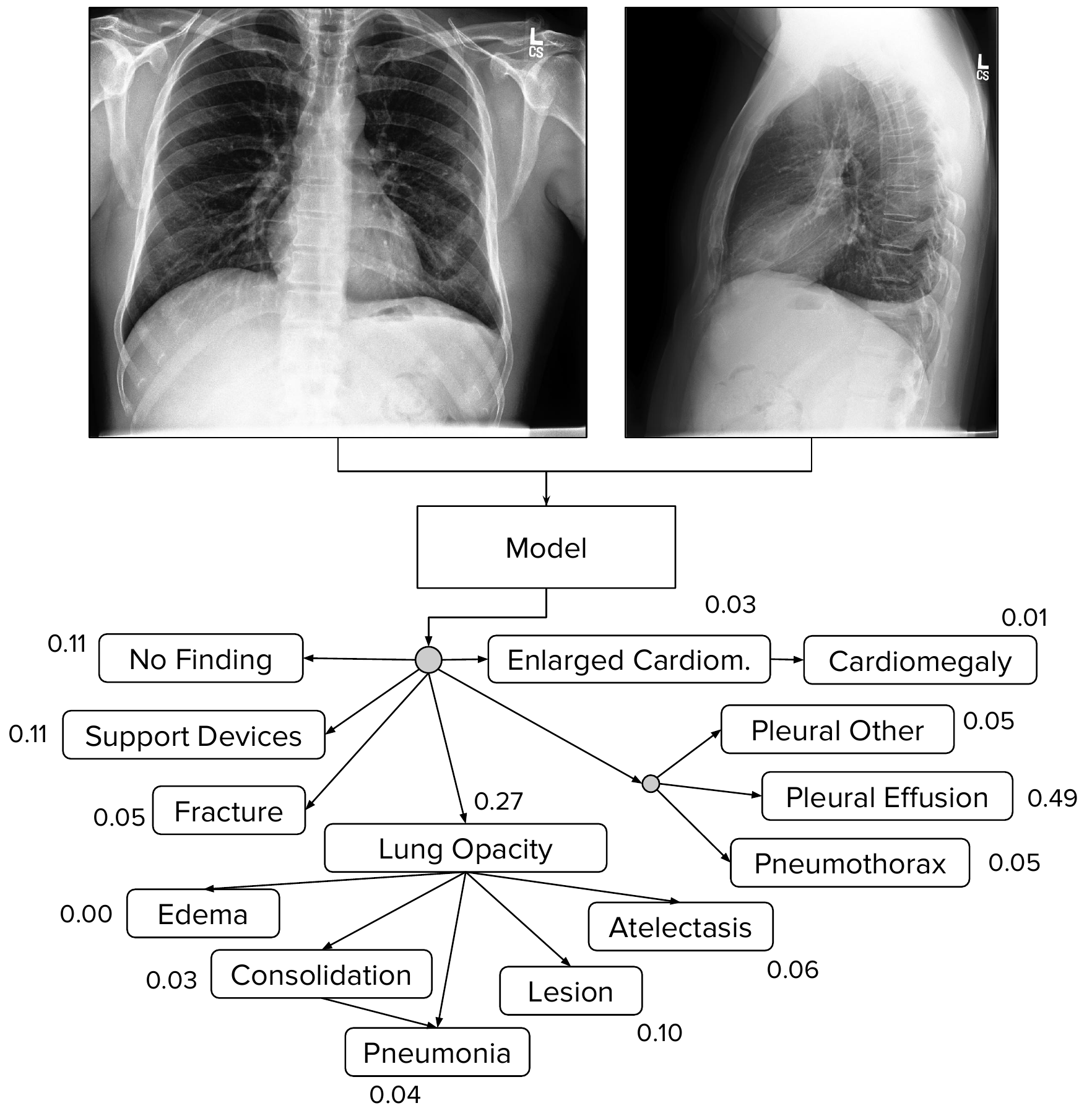}
\caption{The CheXpert task is to predict the probability of different observations from multi-view chest radiographs.}
\label{figure:main}
\end{figure}
\section{Introduction}

Chest radiography is the most common imaging examination globally, critical for screening, diagnosis, and management of many life threatening diseases. Automated chest radiograph interpretation at the level of practicing radiologists could provide substantial benefit in many medical settings, from improved workflow prioritization and clinical decision support to large-scale screening and global population health initiatives. For progress, there is a need for labeled datasets that (1) are large, (2) have strong reference standards, and (3) provide expert human performance metrics for comparison. 


In this work, we present CheXpert (\textbf{Ch}est e\textbf{Xpert}), a large dataset for chest radiograph interpretation. The dataset consists of 224,316 chest radiographs of 65,240 patients labeled for the presence of 14 common chest radiographic observations. We design a labeler that can extract observations from free-text radiology reports and capture uncertainties present in the reports by using an uncertainty label.

The CheXpert task is to predict the probability of 14 different observations from multi-view chest radiographs (see Figure~\ref{figure:main}). We pay particular attention to uncertainty labels in the dataset, and investigate different approaches towards incorporating those labels into the training process. We assess the performance of these uncertainty approaches on a validation set of 200 labeled studies, where ground truth is set by a consensus of 3 radiologists who annotated the set using the radiographs. We evaluate the approaches on 5 observations selected based on their clinical significance and prevalence in the dataset, and find that different uncertainty approaches are useful for different observations.

We compare the performance of our final model to 3 additional board certified radiologists on a test set of 500 studies on which the consensus of 5 separate board-certified radiologists serves as ground truth. We find that on 4 out of 5 pathologies, the model ROC and PR curves lie above at least 2 of 3 radiologist operating points. We make our dataset publicly available to encourage further development of models.
\begin{table}[t]
\centering
\resizebox{\columnwidth}{!}{
\begin{tabular}{lrrr}
\toprule
Pathology &               Positive (\%) &              Uncertain (\%) &               Negative (\%) \\
\midrule
No Finding                 &   16627 (8.86) &        0 (0.0) &  171014 (91.14) \\
Enlarged Cardiom. &    9020 (4.81) &   10148 (5.41) &  168473 (89.78) \\
Cardiomegaly               &  23002 (12.26) &    6597 (3.52) &  158042 (84.23) \\
Lung Lesion                &    6856 (3.65) &    1071 (0.57) &  179714 (95.78) \\
Lung Opacity           &  92669 (49.39) &    4341 (2.31) &    90631 (48.3) \\
Edema                      &  48905 (26.06) &   11571 (6.17) &  127165 (67.77) \\
Consolidation              &   12730 (6.78) &  23976 (12.78) &  150935 (80.44) \\
Pneumonia                  &    4576 (2.44) &   15658 (8.34) &  167407 (89.22) \\
Atelectasis                &  29333 (15.63) &  29377 (15.66) &  128931 (68.71) \\
Pneumothorax               &   17313 (9.23) &    2663 (1.42) &  167665 (89.35) \\
Pleural Effusion           &  75696 (40.34) &    9419 (5.02) &  102526 (54.64) \\
Pleural Other              &     2441 (1.3) &    1771 (0.94) &  183429 (97.76) \\
Fracture                   &    7270 (3.87) &     484 (0.26) &  179887 (95.87) \\
Support Devices            &  105831 (56.4) &     898 (0.48) &   80912 (43.12) \\
\bottomrule
\end{tabular}
}
\caption{The CheXpert dataset consists of 14 labeled observations. We report the number of studies which contain these observations in the training set.}
\label{table:label_prevalences}
\end{table}

\section{Dataset}
CheXpert is a large public dataset for chest radiograph interpretation, consisting of 224,316 chest radiographs of 65,240 patients labeled for the presence of 14 observations as positive, negative, or uncertain. We report the prevalences of the labels for the different obsevations in Table~\ref{table:label_prevalences}.

\subsection{Data Collection and Label Selection}
We retrospectively collected chest radiographic studies from Stanford Hospital, performed between October 2002 and July 2017 in both inpatient and outpatient centers, along with their associated radiology reports.
From these, we sampled a set of 1000 reports for manual review by a board-certified radiologist to determine feasibility for extraction of observations. We decided on 14 observations based on the prevalence in the reports and clinical relevance, conforming to the Fleischner Society's recommended glossary \cite{hansell2008fleischner} whenever applicable.
``Pneumonia'', despite being a clinical diagnosis, was included as a label in order to represent the images that suggested primary infection as the diagnosis. The ``No Finding'' observation was intended to capture the absence of all pathologies.

\begin{figure}[t]
\centering
\includegraphics[width=\columnwidth]{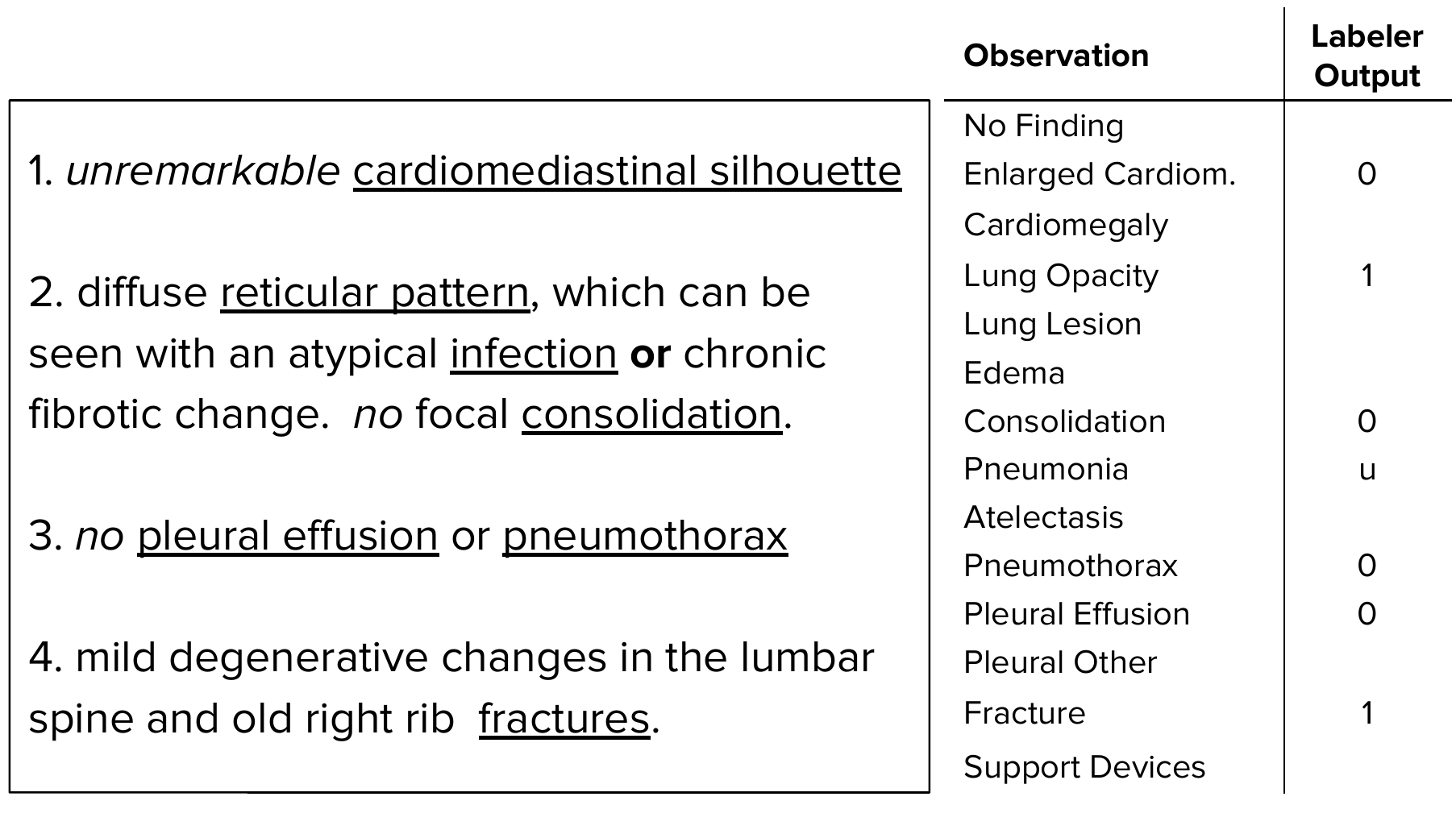}
\caption{Output of the labeler when run on a report sampled from our dataset. In this case, the labeler correctly extracts all of the mentions in the report (underline) and classifies the uncertainties (bolded) and negations (italicized).}
\label{figure:label_example}
\end{figure}

\subsection{Label Extraction from Radiology Reports}
We developed an automated rule-based labeler to extract observations from the free text radiology reports to be used as structured labels for the images. Our labeler is set up in three distinct stages: mention extraction, mention classification, and mention aggregation.

\subsubsection{Mention Extraction}
The labeler extracts mentions from a list of observations from the \textit{Impression} section of radiology reports, which summarizes the key findings in the radiographic study. 
A large list of phrases was manually curated by multiple board-certified radiologists
to match various ways observations are mentioned in the reports.

\subsubsection{Mention Classification}
After extracting mentions of observations, we aim to classify them as negative (``no evidence of pulmonary edema, pleural effusions or pneumothorax''), uncertain (``diffuse reticular pattern may represent mild interstitial pulmonary edema''), or positive (``moderate bilateral effusions and bibasilar opacities''). The `uncertain' label can capture both the uncertainty of a radiologist in the diagnosis 
as well as ambiguity inherent in the report (``heart size is stable'').  The mention classification stage is a 3-phase pipeline consisting of pre-negation uncertainty, negation, and post-negation uncertainty. Each phase consists of rules which are matched against the mention; if a match is found, then the mention is classified accordingly (as uncertain in the first or third phase, and as negative in the second phase). If a mention is not matched in any of the phases, it is classified as positive.

Rules for mention classification are designed on the universal dependency parse of the report. To obtain the universal dependency parse, we follow a procedure similar to \citet{peng2018negbio}: first, the report is split and tokenized into sentences using NLTK \cite{bird2009natural}; then, each sentence is parsed using the Bllip parser trained using David McClosky's biomedical model \cite{charniak2005coarse,mcclosky2010any}; finally, the universal dependency graph of each sentence is computed using Stanford CoreNLP \cite{de2014universal}.

\subsubsection{Mention Aggregation}
We use the classification for each mention of observations to arrive at a final label for 14 observations that consist of 12 pathologies as well as the ``Support Devices'' and ``No Finding'' observations.
Observations with at least one mention that is positively classified in the report is assigned a positive ($1$) label. An observation is assigned an uncertain ($u$) label if it has no positively classified mentions and at least one uncertain mention, and a negative label if there is at least one negatively classified mention. We assign ($blank$) if there is no mention of an observation. The ``No Finding'' observation is assigned a positive label ($1$) if there is no pathology classified as positive or uncertain. An example of the labeling system run on a report is shown in Figure~\ref{figure:label_example}.





\begin{table}[t]
\centering
\resizebox{\columnwidth}{!}{
\begin{tabular}{lrlrlrl}
\toprule
 & \multicolumn{2}{c}{Mention F1} & \multicolumn{2}{c}{Negation F1} & \multicolumn{2}{c}{Uncertain F1}\\
 Category                   &   NIH & Ours   &   NIH & Ours   &   NIH & Ours   \\
\midrule
 Atelectasis                &         0.976 & \textbf{0.998}  &          0.526 & \textbf{0.833}   &           0.661 & \textbf{0.936}    \\
 Cardiomegaly               &         0.647 & \textbf{0.973}  &          0.000     & \textbf{0.909}   &           0.211 & \textbf{0.727}    \\
 Consolidation              &         0.996 & \textbf{0.999}  &          0.879 & \textbf{0.981}   &           0.438 & \textbf{0.924}    \\
 Edema                      &         0.978 & \textbf{0.993}  &          0.873 & \textbf{0.962}   &           0.535 & \textbf{0.796}    \\
 Pleural Effusion           &         0.985 & \textbf{0.996}  &          0.951 & \textbf{0.971}   &           0.553 & \textbf{0.707}    \\
 Pneumonia                  &         0.660  & \textbf{0.992}  &          0.703 & \textbf{0.750}   &           0.250  & \textbf{0.817}    \\
 Pneumothorax               &         0.993 & \textbf{1.000}  &          0.971 & \textbf{0.977}   &           0.167 & \textbf{0.762}    \\
  \midrule
 Enlarged Cardiom. &       N/A     & 0.935          &        N/A     & 0.959           &         N/A     & 0.854            \\
 Lung Lesion                &       N/A     & 0.896          &        N/A     & 0.900           &         N/A     & 0.857            \\
 Lung Opacity           &       N/A     & 0.966          &        N/A     & 0.914           &         N/A     & 0.286            \\
 Pleural Other              &       N/A     & 0.850          &        N/A     & 1.000           &         N/A     & 0.769            \\
 Fracture                   &       N/A     & 0.975          &        N/A     & 0.807           &         N/A     & 0.800            \\
 Support Devices            &       N/A     & 0.933          &        N/A     & 0.720           &         N/A     & N/A              \\
 No Finding                 &       N/A     & 0.769          &        N/A     & N/A             &         N/A     & N/A              \\
 \midrule
 Macro-average              &       N/A     & 0.948          &        N/A     & 0.899           &         N/A     & 0.770            \\
 Micro-average              &       N/A     & 0.969          &        N/A     & 0.952           &         N/A     & 0.848            \\
\bottomrule
\end{tabular}
}
\caption{Performance of the labeler of NIH 
and our labeler on the report evaluation set on tasks of mention extraction, uncertainty detection, and negation detection, as measured by the F1 score. The Macro-average and Micro-average rows are computed over all 14 observations.}
\label{table:label_results}
\end{table}

\section{Labeler Results}
We evaluate the performance of the labeler and compare it to the performance of another automated radiology report labeler on a report evaluation set.

\subsection{Report Evaluation Set}
The report evaluation set consists of 1000 radiology reports from 1000 distinct randomly sampled patients that do not overlap with the patients whose studies were used to develop the labeler. Two board-certified radiologists without access to additional patient information annotated the reports to label whether each observation was mentioned as confidently present (1), confidently absent (0), uncertainly present (u), or not mentioned (blank), after curating a list of labeling conventions to adhere to.
After both radiologists independently labeled each of the 1000 reports, disagreements were resolved by consensus discussion. The resulting annotations serve as ground truth on the report evaluation set.

\subsection{Comparison to NIH labeler}
On the radiology report evaluation set, we compare our labeler against the method employed in \citet{peng2018negbio} which was used to annotate another large dataset of chest radiographs using radiology reports \cite{wang_chestx-ray8:_2017}. We evaluate labeler performance on three tasks: mention extraction, negation detection, and uncertainty detection. For the mention extraction task, we consider any assigned label ($1$, $0$, or $u$) as positive and $blank$ as negative. On the negation detection task, we consider $0$ labels as positive and all other labels as negative. On the uncertainty detection task, we consider $u$ labels as positive and all other labels as negative. We report the F1 scores of the labeling algorithms for each of these tasks.

Table~\ref{table:label_results} shows the performance of the labeling methods. Across all observations and on all tasks, our labeling algorithm achieves a higher F1 score. On negation detection, our labeling algorithm significantly outperforms the NIH labeler on Atelectasis and Cardiomegaly, and achieves notably better performance on Consolidation and Pneumonia. On uncertainty detection, our labeler shows large gains over the NIH labeler, particularly on Cardiomegaly, Pneumonia, and Pneumothorax. 

We note three key differences between our method and the method of \citet{wang_chestx-ray8:_2017}. First, we do not the use automatic mention extractors like MetaMap or DNorm, which we found produced weak extractions when applied to our collection of reports. Second, we incorporate several additional rules in order to capture the large variation in the ways negation and uncertainty are conveyed. Third, we split uncertainty classification of mentions into pre-negation and post-negation, which allowed us to resolve cases of uncertainty rules double matching with negation rules in the reports. For example, the following phrase ``cannot exclude pneumothorax.'' conveys uncertainty in the presence of pneumothorax. Without the pre-negation stage, the `pneumothorax' match is classified as negative due to the `exclude XXX' rule. However, by applying the `cannot exclude' rule in the pre-negation stage, this observation can be correctly classified as uncertain.




\begin{table*}[t]
\centering
\resizebox{\textwidth}{!}{
\begin{tabular}{rlllll}
  \toprule
 & Atelectasis & Cardiomegaly & Consolidation & Edema & Pleural Effusion \\ 
  \midrule
U-Ignore & 0.818 (0.759,0.877) & \textit{0.828 (0.769,0.888)} & 0.938 (0.905,0.970) & 0.934 (0.893,0.975) & 0.928 (0.894,0.962) \\ 
U-Zeros & \textit{0.811 (0.751,0.872)} & 0.840 (0.783,0.897) & 0.932 (0.898,0.966) & 0.929 (0.888,0.970) & 0.931 (0.897,0.965) \\ 
U-Ones & \textbf{0.858 (0.806,0.910)} & 0.832 (0.773,0.890) & 0.899 (0.854,0.944) & 0.941 (0.903,0.980) & 0.934 (0.901,0.967) \\ 
U-SelfTrained & 0.833 (0.776,0.890) & 0.831 (0.770,0.891) & 0.939 (0.908,0.971) & 0.935 (0.896,0.974) & 0.932 (0.899,0.966) \\ 
U-MultiClass & 0.821 (0.763,0.879) & \textbf{0.854 (0.800,0.909)} & 0.937 (0.905,0.969) & 0.928 (0.887,0.968) & 0.936 (0.904,0.967) \\
   \bottomrule
\end{tabular}
}
\caption{AUROC scores on the validation set of the models trained using different approaches to using uncertainty labels. For each of the uncertainty approaches, we choose the best 10 checkpoints per run using the average ROC across the competition tasks. We run each model three times, and take the ensemble of the 30 generated checkpoints on the validation set.}
\label{table:vision_uncertainty}
\end{table*}

\section{Model}
We train models that take as input a single-view chest radiograph and output the probability of each of the 14 observations. When more than one view is available, the models output the maximum probability of the observations across the views.

\subsection{Uncertainty Approaches}
The training labels in the dataset for each observation are either $0$ (negative), $1$ (positive), or $u$ (uncertain). We explore different approaches to using the uncertainty labels during the model training.

\subsubsection{Ignoring} A simple approach to handling uncertainty is to ignore the $u$ labels during training, which serves as a baseline to compare approaches which explicitly incorporate the uncertainty labels. In this approach (called \textit{U-Ignore}), we optimize the sum of the \textit{masked} binary cross-entropy losses over the observations, masking the loss for the observations which are marked as uncertain for the study. Formally, the loss for an example X is given by

\begin{align*}
L(X,y) = -\sum_{o} & \mathbbm{1}\{y_o\neq u\}[y_o \log p(Y_o=1|X) \\
& + (1 - y_o)\log p(Y_o=0|X)],
\end{align*}
where $X$ is the input image, $y$ is the vector of labels of length 14 for the study, and the sum is taken over all 14 observations.
Ignoring the uncertainty label is analogous to the listwise (complete case) deletion method for imputation \cite{graham2009missing}, which is when all cases with a missing value are deleted. Such methods can produce biased models if the cases are not missing completely at random. In this dataset, uncertainty labels are quite prevalent for some observations: for Consolidation, the uncertainty label is almost twice as as prevalent (12.78\%) as the positive label (6.78\%), and thus this approach ignores a large proportion of labels, reducing the effective size of the dataset.

\subsubsection{Binary Mapping}
We investigate whether the uncertain labels for any of the observations can be replaced by the $0$ label or the $1$ label. In this approach, we map all instances of $u$ to $0$ (\textit{U-Zeroes} model), or all to $1$ (\textit{U-Ones} model).

These approaches are similar to zero imputation strategies in statistics, and mimic approaches in multi-label classification methods where missing examples are used as negative labels \cite{kolesov2014multilabel}. If the uncertainty label does convey semantically useful information to the classifier, then we expect that this approach can distort the decision making of classifiers and degrade their performance.


\subsubsection{Self-Training}
One framework for approaching uncertainty labels is to consider them as unlabeled examples, lending its way to semi-supervised learning \cite{zhu2006semi}. Most closely tied to our setting is \textit{multi-label learning with missing labels} (MLML) \cite{wu2015multi}, which aims to handle multi-label classification given training instances that have a partial annotation of their labels.

We investigate a self-training approach (\textit{U-SelfTrained}) for using the uncertainty label. In this approach, we first train a model using the \textit{U-Ignore} approach (that ignores the $u$ labels during training) to convergence, and then use the model to make predictions that re-label each of the uncertainty labels with the probability prediction outputted by the model. We do not replace any instances of $1$ or $0$s. On these relabeled examples, we set up loss as the mean of the binary cross-entropy losses over the observations.

Our work follows the approach of \cite{yarowsky1995unsupervised}, who train a classifier on labeled examples and then predict on unlabeled examples labeling them when the prediction is above a certain threshold, and repeating until convergence. \cite{radosavovic2017data} build upon the self-training technique and remove the need for iteratively training models, predicting on transformed versions of the inputs instead of training multiple models, and output a target label for each unlabeled example; soft labels, which are continuous probability outputs rather than binary, have also been used \cite{hinton2015distilling,li2017learning}.

\subsubsection{3-Class Classification}
We finally investigate treating the $u$ label as its own class, rather than mapping it to a binary label, for each of the 14 observations. We hypothesize that with this approach, we can better incorporate information from the image by supervising uncertainty, allowing the network to find its own representation of uncertainty on different pathologies. In this approach (\textit{U-MultiClass} model), for each observation, we output the probability of each of the 3 possible classes $\{p_0, p_1, p_u\} \in [0, 1]$, $p_0 + p_1 + p_u = 1$. We set up the loss as the mean of the multi-class cross-entropy losses over the observations.
At test time, for the probability of a particular observation, we output the probability of the positive label after applying a softmax restricted to the positive and negative classes.

\subsection{Training Procedure}
We follow the same architecture and training process for each of the uncertainty approaches. We experimented with several convolutional neural network architectures, specifically ResNet152, DenseNet121, Inception-v4, and SE-ResNeXt101, and found that the DenseNet121 architecture produced the best results. Thus we used DenseNet121 for all our experiments. Images are fed into the network with size $320 \times 320$ pixels. We use the Adam optimizer with default $\beta$-parameters of $\beta_1 = 0.9$, $\beta_2 = 0.999$ and learning rate $1\times 10^{-4}$ which is fixed for the duration of the training. Batches are sampled using a fixed batch size of 16 images. We train for 3 epochs, saving checkpoints every 4800 iterations.

\begin{figure*}[t]
\centering
\includegraphics[width=\textwidth]{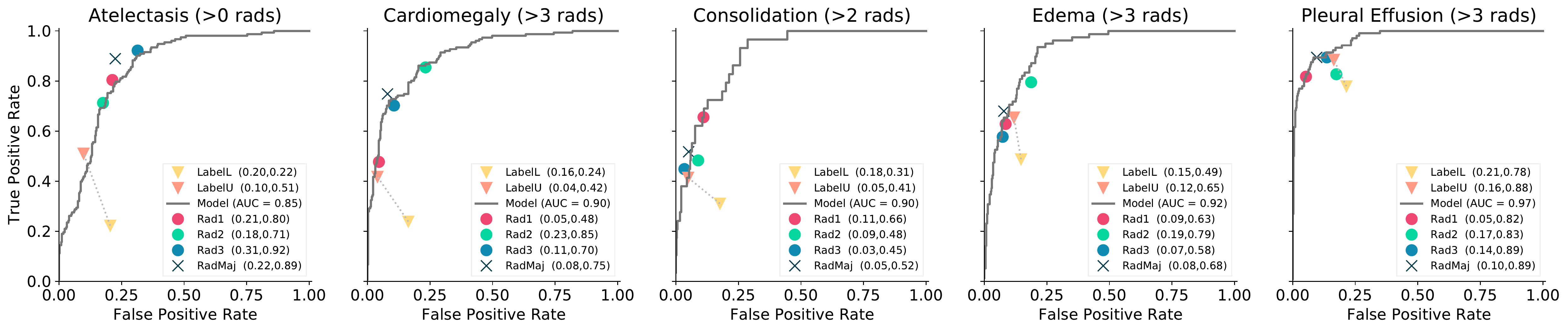}\\
\includegraphics[width=\textwidth]{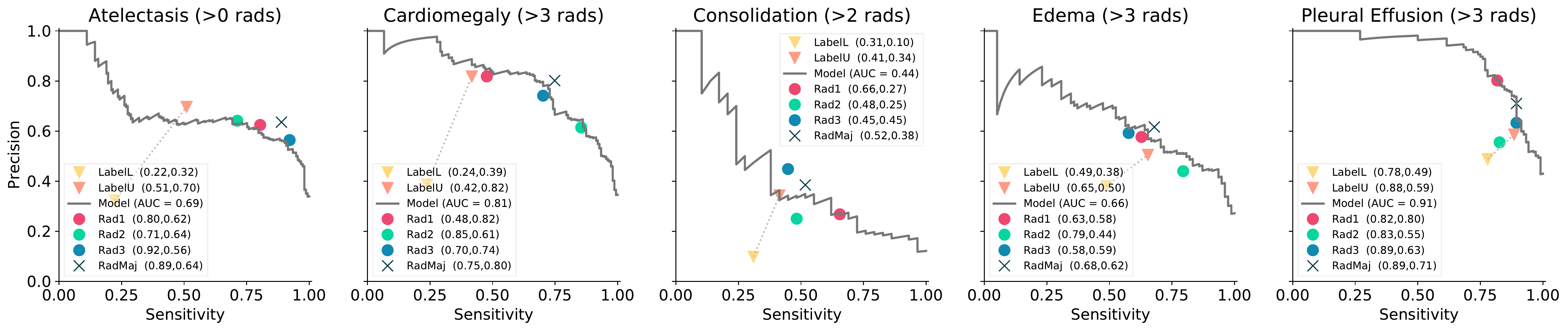}\\
\caption{We compare the performance of 3 radiologists to the model against the test set ground truth in both the ROC and the PR space. We examine whether the radiologist operating points lie below the curves to determine if the model is superior to the radiologists. We also compute the lower (LabelL) and upper bounds (LabelU) of the performance of the labels extracted automatically from the radiology report using our labeling system against the test set ground truth.}
\label{figure:vision_compare_to_rads}
\end{figure*}
\begin{figure*}[t]
\centering

%
%
%
\begin{subfigure}[t]{0.42\textwidth}
\setlength{\tabcolsep}{2pt}
\begin{tabular}{cc}
\centering
\includegraphics[width=0.48\textwidth]{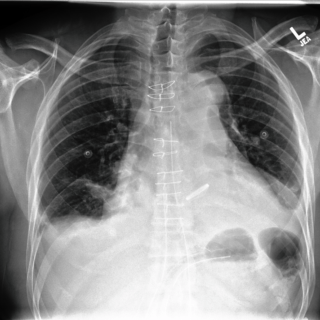} &
\includegraphics[width=0.48\textwidth]{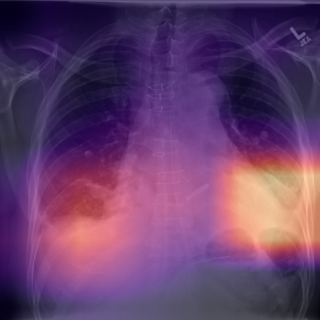}\\
\includegraphics[width=0.48\textwidth]{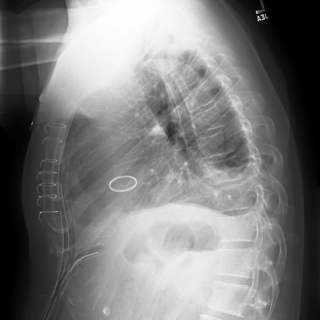} &
\includegraphics[width=0.48\textwidth]{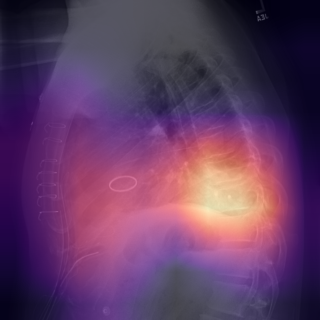}\\
\end{tabular}
\caption{Frontal and lateral radiographs of the chest in a patient with bilateral pleural effusions; the model localizes the effusions on both the frontal (top) and lateral (bottom) views, with predicted probabilities $p=0.936$ and $p=0.939$ on the frontal and lateral views respectively.}
\label{fig:cam_cardiomegaly}
\end{subfigure}%
\setlength{\tabcolsep}{10pt}%
\hspace{3em}
\begin{subfigure}[t]{0.45\textwidth}
\setlength{\tabcolsep}{2pt}
\begin{tabular}{cc}
\centering
\includegraphics[width=0.48\textwidth]{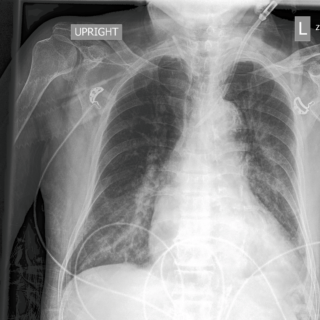} &
\includegraphics[width=0.48\textwidth]{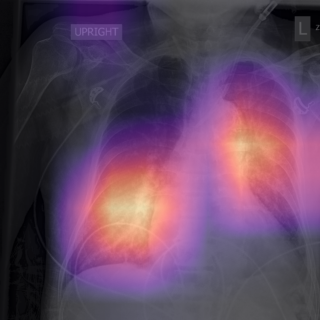}
\end{tabular}
\caption{Single frontal radiograph of the chest demonstrates bilateral mid and lower lung interstitial predominant opacities and cardiomegaly most consistent with cardiogenic pulmonary edema. The model accurately classifies the edema by assigning a probability of $p=0.824$ and correctly localizes the pulmonary edema. Two independent radiologist readers misclassified this examination as negative or uncertain unlikely for edema.}
\label{fig:cam_nodule}
\end{subfigure}%
%
%
%

\caption{The final model localizes findings in radiographs using Gradient-weighted Class Activation Mappings. The interpretation of the radiographs in the subcaptions is provided by a board-certified radiologist.}
\label{figure:vision_cam}
\end{figure*}
\section{Validation Results}
We compare the performance of the different uncertainty approaches on a validation set on which the consensus of radiologist annotations serves as ground truth.

\subsection{Validation Set}
The validation set contains 200 studies from 200 patients randomly sampled from the full dataset with no patient overlap with the report evaluation set. Three board-certified radiologists individually annotated each of the studies in the validation set, classifying each observation into one of present, uncertain likely, uncertain unlikely, and absent. Their annotations were binarized such that all present and uncertain likely cases are treated as positive and all absent and uncertain unlikely cases are treated as negative. The majority vote of these binarized annotations is used to define a strong ground truth \citep{gulshan2016development}.

\subsection{Comparison of Uncertainty Approaches}
\subsubsection{Procedure}
We evaluate the approaches using the area under the receiver operating characteristic curve (AUC) metric.
We focus on the evaluation of 5 observations which we call the competition tasks, selected based of clinical importance and prevalence in the validation set: (a) Atelectasis, (b) Cardiomegaly, (c) Consolidation, (d) Edema, and (e) Pleural Effusion.
We report the 95\% two-sided confidence intervals of the AUC using the non-parametric method by DeLong \cite{delong1988comparing,sun2014fast}. For each pathology, we also test whether the AUC of the best-performing approach is significantly greater than the AUC of the worst-performing approach using the one-sided DeLong’s test for two correlated ROC curves \cite{delong1988comparing}.
We control for multiple hypothesis testing using the Benjamini-Hochberg procedure \cite{benjamini1995controlling}; an adjusted p-value $< 0.05$ indicates statistical significance.

\subsubsection{Model Selection}
For each of the uncertainty approaches, we choose the best 10 checkpoints per run using the average AUC across the competition tasks. We run each model three times, and take the ensemble of the 30 generated checkpoints on the validation set by computing the mean of the output probabilities over the 30 models.

\subsubsection{Results}
The validation AUCs achieved by the different approaches to using the uncertainty labels are shown in Table~\ref{table:vision_uncertainty}.
There are a few significant differences between the performance of the uncertainty approaches. On Atelectasis, the \textit{U-Ones model} (AUC=0.858) significantly outperforms ($p=0.03$) the \textit{U-Zeros model} (AUC=0.811). On Cardiomegaly, we observe that the \textit{U-MultiClass} model (AUC=0.854) performs significantly better ($p<0.01$) than the \textit{U-Ignore} model (AUC=0.828).
On Consolidation, Edema and Pleural Effusion, we do not find the best models to be significantly better than the worst.

\subsubsection{Analysis}
We find that ignoring the uncertainty label is not an effective approach to handling uncertainty in the dataset, and is particularly ineffective on Cardiomegaly. Most of the uncertain Cardiomegaly cases are borderline cases such as ``minimal cardiac enlargement'', which if ignored, would likely cause the model to perform poorly on cases which are difficult to distinguish. However, explicitly supervising the model to distinguish between borderline and non-borderline cases (as in the \textit{U-MultiClass} approach) could enable the model to better disambiguate the borderline cases. Moreover, assignment of the Cardiomegaly label when the heart is mentioned in the impression are difficult to categorize in many cases, particularly for common mentions such as ``unchanged appearance of the heart'' or ``stable cardiac contours'' either of which could be used in both enlarged and non-enlarged cases. These cases were classified as uncertain by the labeler, and therefore the binary assignment of 0s and 1s in this setting fails to achieve optimal performance as there is insufficient information conveyed by these modifications.  

In the detection of Atelectasis, the \textit{U-Ones} approach performs the best, hinting that the uncertainty label for this observation is effectively utilized when treated as positive. We expect that phrases such as ``possible atelectasis'' or ``may be atelectasis,'' were meant to describe the most likely findings in the image, rather than convey uncertainty, which supports the good performance of \textit{U-Ones} on this pathology. We suspect a similar explanation for the high performance of \textit{U-Ones} on Edema, where uncertain phrases like ``possible mild pulmonary edema'' in fact convey likely findings. In contrast, the \textit{U-Ones} approach performs worst on the Consolidation label, whereas the \textit{U-Zeros} approach performs the best.
We also note that Atelectasis and Consolidation are often mentioned together in radiology reports. For example, the phrase ``findings may represent atelectasis versus consolidation'' is very common. In these cases, our labeler assigns uncertain for both observations, but we find that in the ground truth panel review that many of these sorts of uncertainty cases are often instead resolved as Atelectasis-positive and Consolidation-negative.


\section{Test Results}
We compare the performance of our final model to radiologists on a test set. We selected the final model based on the best performing ensemble on each competition task on the validation set: \textit{U-Ones} for Atelectasis and Edema, \textit{U-MultiClass} for Cardiomegaly and Pleural Effusion, and \textit{U-SelfTrained} for Consolidation. 

\subsection{Test Set}\label{test_set}
The test set consists of 500 studies from 500 patients randomly sampled from the 1000 studies in the report test set. Eight board-certified radiologists individually annotated each of the studies in the test set following the same procedure and post-processing as described for the validation set. The majority vote of 5 radiologist annotations serves as a strong ground truth: 3 of these radiologists were the same as those who annotated the validation set and the other 2 were randomly sampled. The remaining 3 radiologist annotations were used to benchmark radiologist performance.

\subsection{Comparison to Radiologists}
\subsubsection{Procedure}
For each of the 3 individual radiologists and for their majority vote, we compute sensitivity (recall), specificity, and precision against the test set ground truth. To compare the model to radiologists, we plot the radiologist operating points with the model on both the ROC and Precision-Recall (PR) space. We examine whether the radiologist operating points lie below the curves to determine if the model is superior to the radiologists.
We also compute the performance of the labels extracted automatically from the radiology report using our labeling system against the test set ground truth. We convert the uncertainty labels to binary labels by computing the upper bound of the labels performance (by assigning the uncertain labels to the ground truth values) and the lower bound of the labels (by assigning the uncertain labels to the opposite of the ground truth values), and plot the two operating points on the curves, denoted \textit{LabelU} and \textit{LabelL} respectively. We also measure calibration of the model before and after applying post-processing calibration techniques, namely isotonic regression \cite{zadrozny2002transforming} and Platt scaling \cite{platt1999probabilistic}, using the scaled Brier score \cite{steyerberg2008clinical}.

\subsubsection{Results}
Figure~\ref{figure:vision_compare_to_rads} illustrates these plots on all competition tasks. The model achieves the best AUC on Pleural Effusion (0.97), and the worst on Atelectasis (0.85). The AUC of all other observations are at least 0.9. The model achieves the best AUPRC on Pleural Effusion (0.91) and the worst on Consolidation (0.44). On Cardiomegaly, Edema, and Pleural Effusion, the model achieves higher performance than all 3 radiologists but not their majority vote. On Consolidation, model performance exceeds 2 of the 3 radiologists, and on Atelectasis, all 3 radiologists perform better than the model. On all competition tasks, the lower bound of the report labels lies below the model curves. On all tasks besides Atelectasis, the upper bound of the report label lies on or below the model operating curves. On most of the tasks, the upper bound of the labeler performs comparably to the radiologists. The average scaled Brier score of the model before post-processing calibration is 0.110, after isotonic regression is 0.107, and after platt scaling is 0.101.

\subsubsection{Limitations} We acknowledge two limitations to performing this comparison. First, neither the radiologists nor the model had access to patient history or previous examinations, which has been shown to decrease diagnostic performance in chest radiograph interpretation \cite{potchen1979effect,berbaum1985effect}. Second, no statistical test was performed to assess whether the difference between the performance of the model and the radiologists is statistically significant. 

\subsection{Visualization}
We visualize the areas of the radiograph which the model predicts to be most indicative of each observation using Gradient-weighted Class Activation Mappings (Grad-CAMs) \cite{selvaraju2016grad}. Grad-CAMs use the gradient of an output class into the final convolutional layer to produce a low resolution map which highlights portions of the image which are important in the detection of the output class. Specifically, we construct the map by using the gradient of the final linear layer as the weights and performing a weighted sum of the final feature maps using those weights. We upscale the resulting map to the dimensions of the original image and overlay the map on the image. Some examples of the Grad-CAMs are illustrated in Figure~\ref{figure:vision_cam}.

\section{Existing Chest Radiograph Datasets}

One of the main obstacles in the development of chest radiograph interpretation models has been the lack of datasets with strong radiologist-annotated groundtruth and expert scores against which researchers can compare their models. There are few chest radiographic imaging datasets that are publicly available, but none of them have test sets with strong ground truth or radiologist performances. The Indiana Network for Patient Care hosts the OpenI dataset \cite{demner2015preparing} consisting of 7,470 frontal-view radiographs and radiology reports which have been labeled with key findings by human annotators . The National Cancer Institute hosts the PLCO Lung dataset \cite{gohagan2000prostate} of chest radiographs obtained during a study on lung cancer screening . The dataset contains 185,421 full resolution images, but due to the nature of the collection process, it is has a low prevalence of clinically important pathologies such as Pneumothorax, Consolidation, Effusion, and Cardiomegaly. The MIMIC-CXR dataset \cite{rubin2018large} has been recently announced but is not yet publicly available.

The most commonly used benchmark for developing chest radiograph interpretation models has been the ChestX-ray14 dataset \cite{wang_chestx-ray8:_2017}. Due to the introduction of this large dataset, substantial progress has been made towards developing automated chest radiograph interpretation models \cite{yao2017learning,rajpurkar_chexnet_2017,li2017thoracic,kumar2018boosted,wang2018tienet,guan2018diagnose,yao2018weakly}. However, using the NIH dataset as a benchmark on which to compare models is problematic as the labels in the test set are extracted from reports using an automatic labeler. The CheXpert dataset that we introduce features radiologist-labeled validation and test sets which serve as strong reference standards, as well as expert scores to allow for robust evaluation of different algorithms.

\section{Conclusion}

We present a large dataset of chest radiographs called CheXpert, which features uncertainty labels and radiologist-labeled reference standard evaluation sets. We investigate a few different approaches to handling uncertainty and validate them on the evaluation sets. On a test set with a strong ground truth, we find that our best model outperforms at least 2 of the 3 radiologists in the detection of 4 clinically relevant pathologies. We hope that the dataset will help development and validation of chest radiograph interpretation models towards improving healthcare access and delivery worldwide.

\section{Acknowledgements}
We would like to thank Luke Oakden-Rayner, Yifan Peng, and Susan C. Weber for their help in this work.

\small 
\bibliography{bibliography}
\bibliographystyle{aaai}

\end{document}